\documentclass[runningheads,a4paper]{llncs}

\usepackage{latexsym}
\usepackage{graphicx}
\usepackage{dsfont}
\usepackage[leqno]{amsmath}
\usepackage[all]{xy}
\usepackage{algorithm}
\usepackage{algorithmic}
\usepackage{amsmath}
\usepackage{amssymb}
\usepackage{subfigure}
\usepackage{wrapfig}
  \newdimen\paravsp  \paravsp=1.3ex
\usepackage{times}

\newcommand{\CL}{\textbf{CL}}
\newcommand{\n}{\textbf{n}}

\newcommand{\Cost}{\textbf{Cost}}
\newcommand{\E}{\textbf{E}}

\newcommand{\entropy}{\textbf{H}}

\newcommand{\A}{\mathcal{A}}
\newcommand{\R}{\mathcal{R}}
\renewcommand{\H}{\mathcal{H}}
\renewcommand{\S}{\mathcal{S}}
\renewcommand{\O}{\mathcal{O}}

\newcommand{\ds}{\displaystyle} \topmargin=-10mm
\def\argmin{\operatornamewithlimits{argmin}}

\def\,{\mskip 3mu} \def\>{\mskip 4mu plus 2mu minus 4mu} \def\;{\mskip 5mu plus 5mu} \def\!{\mskip-3mu}
\def\dispmuskip{\thinmuskip= 3mu plus 0mu minus 2mu \medmuskip=  4mu plus 2mu minus 2mu \thickmuskip=5mu plus 5mu minus 2mu}
\def\textmuskip{\thinmuskip= 0mu                    \medmuskip=  1mu plus 1mu minus 1mu \thickmuskip=2mu plus 3mu minus 1mu}
\textmuskip
\def\beq{\dispmuskip\begin{equation}}    \def\eeq{\end{equation}\textmuskip}
\def\beqn{\dispmuskip\begin{displaymath}}\def\eeqn{\end{displaymath}\textmuskip}
\def\bqa{\dispmuskip\begin{eqnarray}}    \def\eqa{\end{eqnarray}\textmuskip}
\def\bqan{\dispmuskip\begin{eqnarray*}}  \def\eqan{\end{eqnarray*}\textmuskip}

\def\paradot#1{\vspace{\paravsp plus 0.5\paravsp minus 0.5\paravsp}\noindent{\bf\boldmath{#1.}}}

\def\aidx#1{}

\usepackage{url}
\begin{document}
\title{Feature Reinforcement Learning in Practice}

\author{Phuong Nguyen \and Peter Sunehag \and Marcus Hutter}
\institute{Research School of Computer Science, CECS, ANU\\
\path|nmphuong@cecs.anu.edu.au|\\
\path|peter.sunehag, marcus.hutter@{anu.edu.au}|}
\maketitle


\begin{abstract}
Following a recent surge in using history-based methods for
resolving perceptual aliasing in reinforcement learning, we
introduce an algorithm based on the feature reinforcement learning
framework called $\Phi$MDP \cite{MH09c}. To create a practical
algorithm we devise a stochastic search procedure for a class of
context trees based on parallel tempering and a specialized
proposal distribution. We provide the first empirical evaluation for $\Phi$MDP. Our proposed algorithm achieves superior performance to the classical U-tree algorithm \cite{AKM96}
and the recent active-LZ algorithm \cite{Far10}, and is
competitive with MC-AIXI-CTW \cite{VNHUS11} that maintains a
bayesian mixture over all context trees up to a chosen depth. We
are encouraged by our ability to compete with this sophisticated
method using an algorithm that simply picks one single model, and uses
Q-learning on the corresponding MDP. Our $\Phi$MDP algorithm is much simpler, yet
consumes less time and memory. These results show promise for our future work on attacking more complex and larger problems.
\end{abstract}

\section{Introduction}

Reinforcement Learning (RL) \cite{SB98} aims to learn how to succeed
in a task through trial and error. This active research area is well
developed for environments that are Markov Decision Processes
(MDPs); however, real world environments are often partially
observable and non-Markovian. The recently introduced Feature Markov Decision Process ($\Phi$MDP) framework \cite{MH09c} attempts to
reduce actual RL tasks to MDPs for the purpose of attacking the
general RL problem where the environment's model as well as the set
of states are unknown. In \cite{SH10}, Sunehag and Hutter take
a step further in the theoretical investigation of Feature
Reinforcement Learning by proving consistency results. In this article,
we develop an actual Feature Reinforcement Learning algorithm and
empirically analyze its performance in a number of environments.

One of the most useful classes of maps ($\Phi$s) that can be used to
summarize histories as states of an MDP, is the class of context
trees. Our stochastic search procedure, the principal component of
our $\Phi$MDP algorithm GS$\Phi$A, works on a subset of
all context trees, called Markov trees. Markov trees have
previously been studied in \cite{Ris83} but under names like
FSMX sources or FSM closed tree sources. The stochastic search
procedure employed for our empirical investigation utilizes a parallel tempering methodology
\cite{CJG91}, \cite{HN96} together with a specialized proposal
distribution. In the experimental section, the performance of the $\Phi$MDP algorithm where stochastic search is conducted over the space of context-tree maps is shown and compared with three other related context tree-based methods. 


Our $\Phi$MDP algorithm is briefly summarized  as follows. First,
perform a certain number of random actions, then use this history to
find a high-quality map by minimizing a cost function that evaluates
the quality of each map. The quality here refers to the ability to
predict rewards using the created states. We perform a search
procedure for uncovering high-quality maps followed by executing
$Q$-learning on the MDP whose states are induced by the detected
optimal map. The current history is then updated with the additional experiences obtained from the interactions with the environment through Q-Learning. After that, we may repeat the procedure but without the random actions.
The repetition refines the current ``optimal'' map, as longer histories provide more useful
information for map evaluation. The ultimate optimal policy of the algorithm is retrieved from the action values Q on the resulting MDP induced from the final optimal map.

\paradot{Contributions}
Our contributions are: extending the original $\Phi$MDP cost
function presented in \cite{MH09c} to allow for more discriminative
learning and more efficient minimization (through stochastic search)
of the cost; identifying the Markov action-observation context trees as
an important class of feature maps for $\Phi$MDP; proposing the
GS$\Phi$A algorithm where several chosen learning and search
procedures are logically combined; providing the first empirical analysis of the $\Phi$MDP
model; and designing a specialized proposal distribution for stochastic
search over the space of Markov trees, which is of critical
importance for finding the best possible $\Phi$MDP agent.


\paradot{Related Work}
Our algorithm is a history-based method. This means that we are
utilizing memory that in principle can be long, but in most of this
article and in the related works is near term. Given a history $h_t$
of observations, actions and rewards we define states
$s_t=\Phi(h_t)$ based on some map $\Phi$. The main class of maps that we will
consider are based on context trees. The classical algorithm of this
sort is U-tree \cite{AKM96}, which uses a local criterion based on a
statistical test for splitting nodes in a context tree; while
$\Phi$MDP employs a global cost function. Because of this advantage,
$\Phi$MDP can potentially be used in conjunction with any optimization
methods to find the optimal model.

There has been a recent surge of interest in history based methods
with the introduction of the active-LZ algorithm \cite{Far10}, which
generalizes the widely used Lempel-Ziv compression scheme to the
reinforcement learning setting and assumes $n$-Markov models of environments; and MC-AIXI-CTW \cite{VNHUS11},
which uses a Bayesian mixture of context trees and incorporates both
the Context Tree Weighting algorithm \cite{WST95} as well as UCT
Monte Carlo planning \cite{KS06}. These can all be viewed as
attempts at resolving perceptual aliasing problems with the help of
short-term memory. This has turned out to be a more tractable
approach than Baum-Welch methods for learning a Partially Observable
Markov Decision Process (POMDP) \cite{Chr92} or Predictive State
Representations \cite{SJR04}. The history based methods attempt to
directly learn the environment states, thereby avoiding the
POMDP-learning problem \cite{KLC98}, \cite{MHC03} which is extremely
hard to solve. Model minimization \cite{GDG03} is a line of works that also seek for a minimal representation of the state space, but focus on solving Markovian problems while $\Phi$MDP and other aforementioned history-based methods target non-Markovian ones. It is also worthy to note that there are various other attempts to find compact representations of MDP state spaces \cite{LWL06}; most of which, unlike our approach, address the planning problem where the MDP model is given


\paradot{Paper Organization}
The paper is organized as follows. Section \ref{sec:prel} introduces
preliminaries on Reinforcement Learning, Markov Decision Processes,
Stochastic Search methods and Context Trees. These are the
components from which the $\Phi$MDP algorithm (GS$\Phi$A) is built.
In Section \ref{sec:agent} we put all of the components into our
$\Phi$MDP algorithm and also describe our specialized search
proposal distribution in detail. Section \ref{sec:exp} presents
experimental results on four domains. Finally Section
\ref{sec:concl} summarizes the main results of this paper, and
briefly suggests possible research directions.



\section {Preliminaries}\label{sec:prel}

\subsection{Markov Decision Processes (MDP)}
An environment is a process which at any discrete time $t$, given action
$a_t\in\mathcal{A}$ produces an observation $o_t\in\mathcal{O}$ and
a corresponding reward $r_t\in\mathbb{R}$. When the process is a
Markov Decision Process \cite{SB98}; $o_t$ represents the
environment state, and hence is denoted by $s_t$ instead. Formally,
a finite MDP is denoted by a quadruple
$\langle\mathcal{S},\mathcal{A},\mathcal{T},\mathcal{R}\rangle$ in
which $\mathcal{S}$ is a finite set of states; $\mathcal{A}$  is a
finite set of actions; $T = (T_{ss'}^a:s,s'\in \mathcal{S},\ a \in
\mathcal{A})$ is a collection of transition probabilities of the next
state $s_{t+1}=s'$ given the current state $s_t = s$ and action
$a_t=a$; and $R = (R_{ss'}^a:s,s'\in \mathcal{S},\ a \in \mathcal{A})$
is a reward function $R_{ss'}^a=\E[r_{t+1}|s_t=s, a_t=a,s_{t+1} = s']$.
The return at time step $t$ is the total discounted reward $R_t =
r_{t+1}+ \gamma{}r_{t+2} + \gamma^2r_{t+3}+ \ldots $, where $\gamma$
is the geometric discount factor ($0 \leq\gamma<1$).

Similarly, the action value in state $s$ following policy $\pi$ is
defined as $Q^{\pi}(s,a) = \E_\pi[R_t|s_t=s, a_t = a] =
\E_{\pi}[\sum^{\infty}_{k=0}\gamma^k{}r_{t+k+1}|s_t = s,
a_t=a]$. For a known MDP, a useful way to find an estimate of the optimal action values
$Q^*$ is to employ the Action-Value Iteration (AVI) algorithm, which
is based on the optimal action-value Bellman equation \cite{SB98},
and iterates the update $Q(s,a) \leftarrow
\sum_{s'}T_{ss'}^a[R_{ss'}^a
    + \gamma\max_{a'}Q(s',a')].$

%

If the MDP model is unknown, an effective estimation technique is
provided by $Q$-learning, which incrementally updates
estimates $Q_t$ through the equation
$$Q(s_t,a_t)\leftarrow Q(s_t,a_t)+\alpha_t(s_t,a_t)err_t$$
where the feedback error $err_t=r_{t+1}+\gamma \max_a
Q(s_{t+1},a)-Q(s_t,a_t)$, and $\alpha_t(s_t,a_t)$ is the learning rate
at time $t$. Under the assumption of sufficient visits of all
state-action pairs, Q-Learning converges if and only if some
conditions of the learning rates are met \cite{BT96}, \cite{SB98}. In
practice a small constant value of the learning rates
($\alpha(s_t,a_t)=\eta$) is, however, often adequate to get a good estimate of
$Q^*$. Q-Learning is off-policy; it directly approximates $Q^*$
regardless of what actions are actually taken. This approach is particularly
beneficial when handling the exploration-exploitation tradeoff in RL.

It is well known that learning by taking greedy actions retrieved from the current estimate $\widehat{Q}$ of $Q^*$ to explore the state-action space generally leads to suboptimal
behavior. The simplest remedy for this inefficiency is to employ the
$\epsilon$-greedy scheme, where with probability $\epsilon>0$ we
take a random action, and with probability $1-\epsilon$ the greedy
action is selected. This method is simple, but has shown to fail to
properly resolve the exploration-exploitation tradeoff. A more
systematic strategy for exploring the unseen scenarios, instead of just
taking random actions, is to use optimistic initial values
\cite{SB98}, \cite{BT02}. To apply this idea to $Q$-Learning, we
simply initialize $Q(s,a)$ with large values. Suppose $R_{\max}$ is
the maximal reward, $Q$ initializations of at least
$\frac{R_{\max}}{1-\gamma}$ are optimistic as
$Q(s,a)\leq\frac{R_{\max}}{1-\gamma}$.

\subsection{Feature Reinforcement Learning}

\paradot{Problem description}
An RL agent aims to find the optimal policy $\pi$ for taking action
$a_t$ given the history of past observations, rewards and actions $h_t =
o_1r_1a_1\ldots{}o_{t-1}r_{t-1}a_{t-1}o_tr_t$ in order to maximize
the long-term reward signal. If the problem satisfies an MDP; as can
be seen above, efficient solutions are available. We aim to attack
the most challenging RL problem where the environment's states and
model are both unknown. In \cite{MH05}, this problem is named the
Universal Artificial Intelligence (AI) problem since almost all AI
problems can be reduced to it.

\paradot{$\Phi$MDP framework}
In \cite{MH09c}, Hutter proposes a history-based method, a
general statistical and information theoretic framework called
$\Phi$MDP. This approach offers a critical preliminary reduction
step to facilitate the agent's ultimate search for the optimal
policy. The general $\Phi$MDP framework endeavors to extract
relevant features for reward prediction from the past history $h_t$
by using a feature map $\Phi$: $\H \rightarrow \S$, where $\H$ is
the set of all finite histories. More specifically, we want the
states $s_t = \Phi(h_t)$ and the resulting tuple $\langle{}\S, \ \A,
\ \R\rangle$ to satisfy the Markov property of an MDP. As
aforementioned, one of the most useful classes of $\Phi$s is the
class of context trees, where each tree maps a history to a single
state represented by the tree itself. A more general class of $\Phi$
is Probabilistic-Deterministic Finite Automata (PDFA)
\cite{VTHCC05}, which map histories to the MDP states where the next
state can be determined from the current state and the next observation.
The primary purpose of $\Phi$MDP is to find a map $\Phi$ so that
rewards of the MDP induced from the map can be predicted well. This
enables us to use MDP solvers, like AVI and Q-learning, on the induced MDP to find
a good policy. The reduction quality of each $\Phi$ is dictated by
the capability of predicting rewards of the resulting MDP induced
from that $\Phi$. A suitable cost function that measures the utility of
$\Phi$s for this purpose is essential, and the optimal $\Phi$ is the
one that minimizes this cost function.

\paradot{Cost function}
The cost used in this paper is an extended version of the original
cost introduced in \cite{MH09c}. We define a cost that measures the
reward predictability of each $\Phi$, or more specifically of the
resulting MDP induced from that $\Phi$. Based on this, our cost
includes the description length of rewards; however, rewards depend
on states as well, so the description length of states must be also added
to the cost. In other words, the cost comprises coding of the
rewards and resulting states, and is defined as follows:

$$\Cost_{\alpha}(\Phi|h_n) :=\alpha \CL(s_{1:n}|a_{1:n})+(1-\alpha)
\CL(r_{1:n}|s_{1:n},a_{1:n})$$ where $s_{1:n}=s_1,...,s_n$ and
$a_{1:n}=a_1,...,a_n$ and $s_t = \Phi(h_t)$ and $h_t = ora_{1:t-1}r_t$
and $0\leq \alpha \leq 1$.  For coding we use the two-part code \cite{WD99}, \cite{PDG07}, hence the code length (CL) is $\CL(x)=\CL(x|\theta)+\CL(\theta)$ where $x$
denotes the data sampled from the model specified by parameters
$\theta$. We employ the optimal codes \cite{CT91} for describing
data $\CL(x|\theta)=\log(1/Pr_\theta(x))$, while parameters are
uniformly encoded to precision $1/\sqrt{\ell(x)}$ where $\ell(x)$ is the sequence length
of $x$ \cite{PDG07}: $\CL(\theta)=\frac{m-1}{2}\log{\ell(x)}$, here $m$ is the number of
parameters. The optimal $\Phi$ is found via the optimization problem
$\Phi^{optimal} = \argmin_{\Phi} \Cost_{\alpha}(\Phi|h_n)$.

Denote $\n_\bullet:=[n_1\ n_2 \ldots\ n_l]$ ($l$ is determined in
specific context); $n_+:=\sum_jn_j$ ($n_j$s are components of vector
$\n_\bullet$); $|\bullet|$ cardinality of a set; $n_{ss'}^{ar'}
:=|\{t:(s_t,a_t,s_{t+1}, r_{t+1})=(s,a,s',r'), \ 1\leq{}t\leq{}n\}|$; and
$\entropy(\textbf{p})=-\sum_{i=1}^lp_i\log{}p_i$ Shannon entropy of a random variable with
distribution $\textbf{p} = [p_1\ p_2 \text{\ldots}\ p_l]$ where
$\sum_{i=1}^l{}p_i = 1$. The state and reward cost functions can, then, be
analytically computed as follows:
\begin{eqnarray*}
    &&\CL(s_{1:n}|a_{1:n}) = \sum_{s, a}\CL(\n^{a+}_{s\bullet}) = \ds\sum_{s,
                    a}n^{a+}_{s+}\entropy\left(\frac{\n^{a+}_{s\bullet}}{n^{a+}_{s+}}\right)
                    + \frac{|\S| -1}{2}\log{n^{a+}_{s+}}\\
    &&\CL(r_{1:n}|s_{1:n},a_{1:n}) = \sum_{s, a, s'}\CL(\n^{a\bullet}_{ss'}) = \sum_{s,a,s'}n^{a+}_{ss'}\entropy\left(\frac{\n^{a\bullet}_{ss'}}{n^{a+}_{ss'}}\right)
                    + \frac{|\R| -1}{2}\log{}n^{a+}_{ss'}
\end{eqnarray*}

As we primarily want to find a $\Phi$ that has the best reward
predictability, the introduction of $\alpha$ is primarily to stress
on reward coding, making costs for high-quality $\Phi$s much lower
with very small $\alpha$ values. In other words, $\alpha$ amplifies
the differences among high-quality $\Phi$s and bad ones; and this
accelerates our stochastic search process described below.

We furthermore replace $\CL(x)$ with $\CL_\beta(x)=\CL(x|\theta)+\beta
\CL(\theta)$ in $\Cost_\alpha$ to define $\Cost_{\alpha,\beta}$ for
the purpose of being able to select the right model given limited
data. The motivation to introduce $\beta$ is the following. For
stationary environments the cost function is analytically of this
form $C_1\times u(\alpha)\times O(n) + C_2\times v(\alpha)\times
t(\beta)\times O(\log(n))$ where $C_1, C_2$ are constants, and $u, v,
t$ are linear functions. The optimal $\Phi$ should be the one with
the smallest value of $C_1\times u(\alpha)$, however, the curse here
is that in practice $C_2\times v(\alpha)$  is often big, so in order
to obtain the optimal $\Phi$ with limited data, a small value of
$\beta$ will help. We assert that with a very large number of
samples $n$, $\alpha$ and $\beta$ can be ignored in the above cost
function (use $\alpha=0.5,\ \beta=1$ as the cost in \cite{MH09c}).
The choice of small $\alpha$ and $\beta$ helps us more quickly to overcome the model penalty and find the optimal map. This strategy is a quite common practice in statistics, and even in the Minimum Description Length
(MDL) community \cite{PDG07}. For instance, AIC \cite{HA74} uses a very small $\beta = 2/\log{}n$.

The interested reader is referred to \cite{MH09c} for more detailed
analytical formulas, and \cite{SH10} for further motivation and
consistency proofs of the $\Phi$MDP model.

\subsection{Context Trees} \label{sec:contexttrees}

The class of maps that we will base our algorithm on is a class of
context trees.

\paradot{Observation Context Tree (OCT)}
OCT is a class of maps $\Phi$ used to extract relevant information
from histories that include only past observations, not actions and
rewards. The presentation of OCT is mainly to facilitate the definitions of the below Action-Observation Context Tree.

\paradot{Definition}
Given an $|\O|$-ary alphabet $\O = \{o^1, o^2, \ldots, o^{|\O|}\}$,
an OCT constructed from the alphabet $\mathcal{O}$ is defined as a $|\O|$-ary tree in which edges coming from any internal node are labeled by letters in $\mathcal{O}$ from left to right in the
order given.

Given an OCT $\mathcal{T}$ constructed from the alphabet $\O$, the state suffix set, or briefly
state set $\mathcal{S} = \{s^1, s^2, \ldots, s^m\} \subseteq \O^*$
induced from $\mathcal{T}$ is defined as the set of all possible
strings of edge labels forming along a path from a leaf node to the
root node of $\mathcal{T}$. $\mathcal{T}$ is called a Markov
tree if it has the so-called Markov property for its associated
state set, that is, for every $s^i \in \S$ and $o^k\in \O$,
$s^io^k$ has a unique suffix $s^j \in \S $. The state set of a
Markov OCT is called Markov state set. OCTs that do not have the
Markov property are identified as non-Markov OCTs. Non-Markov state sets are similarly defined.

\begin{figure}[!h]\centering
\subfigure[\textbf{Binary context trees}] {
    \includegraphics[scale=0.59]{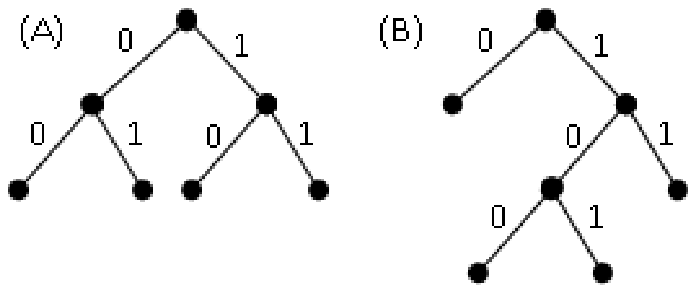}
        \label{fig:binarycontexttrees}
}
\subfigure[\textbf{Trinary context trees}]{
    \includegraphics[scale=0.59]{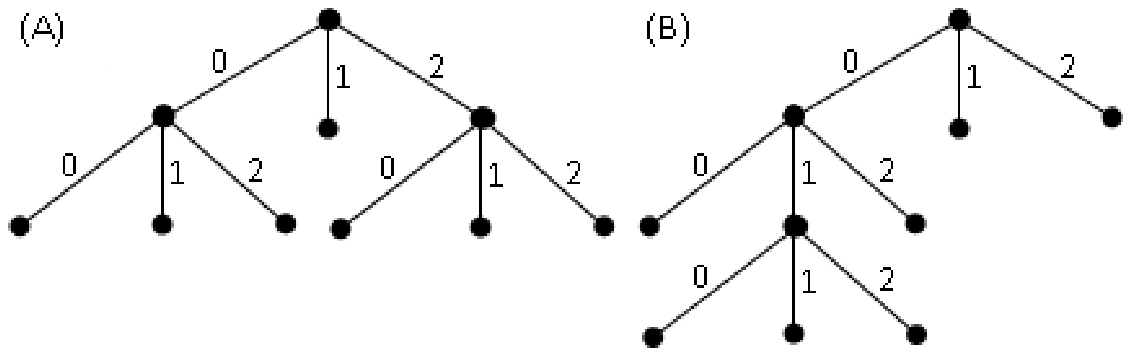}
    \label{fig:trinarycontexttrees}
    }
    \caption{\textbf{Context Trees}}
    \label{fig:contexttrees}
\end{figure}

\paradot{Example}
Figure \ref{fig:binarycontexttrees}(A) and \ref{fig:binarycontexttrees}(B)
respectively represent two binary OCTs of depths two and
three; also Figures \ref{fig:trinarycontexttrees}(A) and
\ref{fig:trinarycontexttrees}(B) illustrate two ternary OCTs of
depths two and three.

As can be seen from Figure \ref{fig:contexttrees}, trees
\ref{fig:binarycontexttrees}(A) and \ref{fig:trinarycontexttrees}(A) are
Markov; on the other hand, trees \ref{fig:binarycontexttrees}(B) and
\ref{fig:trinarycontexttrees}(B) are non-Markov. The state set of tree
\ref{fig:binarycontexttrees}(A) is $\mathcal{S}^{(a)(A)} = \{00, 01, 01,
11\}$; and furthermore with any further observation $o \in
\mathcal{O}$ and $s \in \mathcal{S}^{(a)(A)}$, there exists a unique $s' \in
\mathcal{S}$ which is a suffix of $so$. Hence, tree
\ref{fig:binarycontexttrees}(A) is Markov. Table 1(a) 
 represents the
deterministic relation between $s, \ o$ and $s'$.

\begin{table}[!h]\centering
\subtable[\textbf{Markov property of $\mathcal{S}^{(a)(A)}$}]{\centering
\begin{tabular}{ccc||p{0.45cm}p{0.45cm}p{0.45cm}p{0.45cm}|p{0.45cm}p{0.45cm}p{0.45cm}p{0.45cm}ccc}
  &&$s$ & 00 & 01 & 10 & 11 & 00 & 01 & 10 & 11& &&\\
  \hline
  &&$o$ &     \multicolumn{4}{c|}{0}  &      \multicolumn{4}{c}{1} &&&  \\
  \hline
  &&$s'$ & 00 & 10 & 00 & 10 & 01 & 11 & 01 & 11 &&&
\end{tabular}
\label{tab:markovsuffixset}
}
\subtable[\textbf{Non-markov property of  $\mathcal{S}^{(a)(B)}$}]{\centering
\begin{tabular}{cc||cccc|cccccc}
  &$s$ & 0 & 001 & 101 & 11 & 0 & 001 & 101 & 11 &&\\
  \hline
  &$o$ &     \multicolumn{4}{c|}{0}  &      \multicolumn{4}{c}{1} &&  \\
  \hline
  &$s'$ & 0 & 0 & 0 & 0 & 101 or 001 & 11 & 11 & 11 &&
\end{tabular}
\label{tab:nonmarkovsuffixset}
}
\label{tab:markovproperty}
\caption{\textbf{Markov and Non-Markov properties}}
\end{table}

However, there is no such relation in tree
\ref{fig:binarycontexttrees}(B), or state set $\mathcal{S}^{(a)(B)} = \{0,
001, 101, 11\}$; for $s=0$ and $o=1$, it is ambiguous whether $s'=$101 or 001. Table 1(b) 
clarifies the
non-Markov property of tree \ref{fig:binarycontexttrees}(B).

Similar arguments can be applied for trees \ref{fig:trinarycontexttrees}(A)
and \ref{fig:trinarycontexttrees}(B) to identify their Markov property.

It is also worthy to illustrate how an OCT can be used as a map. We illustrate the mapping using again the OCTs in Figure \ref{fig:contexttrees}. Given two histories including only past observations $h_5=11101$ and $h'_6=211210$, then $\Phi^{(a)(A)}(h_5) = 01, \Phi^{(a)(B)}(h_5) = 101, \Phi^{(b)(A)}(h'_6)=10, \ \text{and}\ \Phi^{(b)(B)}(h'_6)=210$.


\paradot{Action-Observation Context Tree (AOCT)}
AOCTs  are extended from the OCTs presented above for the generic RL
problem where relevant histories contain both actions and observations.

\paradot{Definition}
Given two alphabets, $\O = \{o^1, o^2, \ldots, o^{|\O|}\}$ named
observation set, and $\A = \{a^1, a^2, \ldots, a^{|\A|}\}$ named
action set, an AOCT constructed from the two alphabets
is defined as a tree where any internal node at even depths has
branching factor $|\O|$, and edges coming from such nodes are
labeled by letters in $\O$ from left to right in the order given;
and similarly any internal node at odd depths has branching factor
$|\A|$, and edges coming from these nodes are labeled by letters in
$\A$ also from left to right in the specified order.

The definitions of Markov and non-Markov AOCTs are
similar to those of OCTs except that a next observation is
now replaced by the next action and observation. Formally,  suppose $\mathcal{T}$ is an AOCT
constructed from the above two alphabets; and $\S = \{s^1, s^2,
\ldots, s^m\} \subseteq (\A\times\O)^* \cup \A\times(\A\times\O)^*$ is the state suffix set of
the tree, then $\mathcal{T}$ is defined as a Markov AOCT if it has the Markov property, that is, for every
$1\leq{}i\leq{}m$, $1\leq j\leq |\A|$, and $1\leq k\leq |\O|$ there
exist a unique $1 \leq l \leq m$ such that $s^{l}$ is a suffix
of $s^ia^jo^k$. AOCTs that do not have Markov
property are categorized as non-Markov AOCTs.

The total number of AOCTs up to a certain depth $d$, $K(d)$, can be recursively computed via the
formula $K(d+2) = \{[K(d)]^{|\A|}+1\}^{|\O|}+1$ where $K(0) = 1,
K(1)=2$. As can be easily seen from the recursive formula, the total number of AOCTs is
doubly exponential in the tree depth.

An important point to note here is that in our four experiments presented in Section \ref{sec:exp}, the $\Phi$ space is limited to Markov AOCTs, since as explained above, the state suffix set induced from a non-Markov AOCT does not represent an MDP state set; to put it more clearly, in non-Markov AOCTs, from the next action and observation, we
cannot derive the next state from the current one. The Markov constraint on AOCTs significantly reduces the search space for our stochastic search algorithm. In the U-tree algorithm \cite{AKM96}, no distinction of Marov and non-Markov trees is identified; the algorithm attempts to search for the optimal tree over the whole space of AOCTs.
\subsection{Stochastic search}\label{sec:stochasticsearch}
While we have defined the cost criterion for evaluating
maps, the problem of finding the optimal map remains. When the $\Phi$ space is huge, e.g. context-tree map space where the number of $\Phi$s grows doubly exponentially with the tree depth,
exhaustive search is unable to deal with domains where the optimal $\Phi$ is non-trivial. Stochastic search is a powerful tool for
solving optimization problems where the landscape of the objective
function is complex, and it appears impossible to analytically or
numerically find the exact or even approximate global optimal
solution. A typical stochastic search algorithm starts with a
predefined or arbitrary configuration (initial argument of the
objective function or state of a system), and from this generates a
sequence of configurations based on some predefined
probabilistic criterion; the configuration with the best objective
value will be retained. There are a wide range of stochastic search
methods proposed in the literature \cite{SK06a}; the most popular
among these are simulated-annealing-type algorithms \cite{JSL01},
\cite{SK06}. An essential element of a simulated-annealing (SA) algorithm is a Markov Chain Monte Carlo (MCMC) sampling scheme where a proposed new configuration
$\tilde{y}$ is drawn from a proposal distribution $q(\tilde{y}|y)$, and we then change from configuration $y$ to $\tilde{y}$ with probability $\min\{1,\frac{\pi_{T}(y)q(y|\widetilde{y})}{\pi_{T}(\widetilde{y})q(\widetilde{y}|y)}\}$
where $\pi_T$ is a target distribution. In a simulated-annealing (SA) algorithm where the traditional Metropolis-Hasting sampling scheme is utilized, $\pi_T$ is proportional to $e^{-f(x)/T}$ if $f$ is an objective function that we want to minimize, and $T$ is some positive constant temperature. $\frac{q(y|\widetilde{y})}{q(\widetilde{y}|y)}$ is called the correction factor; it is there to compensate for bias in $q$.

The traditional SA uses an MCMC scheme with some temperature-decreasing strategy.  Although shown to be able to find the global optimum asymptotically \cite{GKJ94}, it generally works
badly in practice as we do not know which temperature cooling scheme
is appropriate for the problem under consideration. Fortunately in the $\Phi$MDP cost function we know typical cost differences between two $\Phi$s
($C\beta\times\log(n)$), so the range of
appropriate temperatures can be significantly reduced. The search process may be improved if we
run a number of SA procedures with various different temperatures.
Parallel Tempering (PT) \cite{CJG91}, \cite{HN96}, an interesting
variant of the traditional SA, significantly improves this
stochastic search process by smartly offering a swapping step,
letting the search procedure use small temperatures for exploitation
and big ones for exploration.

\paradot{Parallel tempering}
PT performs stochastic search over the product space $\mathcal{X}_1\times
\ldots\ \ \times \mathcal{X}_I (\mathcal{X}_i = \mathcal{X}\ \forall
1\leq i\leq I)$, where $\mathcal{X}$ is the objective function's
domain, and $I$ is the parallel factor. Fixed temperatures $T_i$ ($i = 1,
\ldots\ , I$, and $1 < T_1 <T_2<\ldots<T_I$) are chosen for spaces $\mathcal{X}_i$ $(i = 1, \ldots\ , I)$. Temperatures
$T_i$ ($i=1,\ldots,I$) are selected based on the following formula
    $(\frac{1}{T_i}-\frac{1}{T_{i+1}})|\Delta{}H|\approx-\log{}p_a$
where $\Delta{}H$ is the ``typical'' difference between function
values of two successive configurations; and $p_a$ is the lower bound for the swapping acceptance rate. The main steps of each PT
loop are as follows:

\begin{itemize}
    \item $(x^{(t)}_1,\ldots \ ,x^{(t)}_I)$ is the current sampling; draw $u \sim$ Uniform[0,1]
    \item If $u \leq \alpha_0$, update every $x^{(t)}_i$ to $x^{(t+1)}_i$ via some Markov Chain Monte Carlo (MCMC) scheme like Metropolis-Hasting (Parallel step)
    \item If $u > \alpha_0$, randomly choose a neighbor pair, say $i$ and $i+1$, and accept the swap of $x^{(t)}_i$ and $x^{(t)}_{i+1}$ with probability $
            \min\{1, \frac{\pi_{T_i}(x^{(t)}_{i+1})\pi_{T_{i+1}}(x^{(t)}_{i})}{\pi_{T_i}(x^{(t)}_{i})\pi_{T_{i+1}}(x^{(t)}_{i+1})}\}$ (Swapping step).
\end{itemize}

The full details of PT are given in Algorithm \ref{alg:pt}.\newpage

\begin{algorithm}[h!]
\caption{Parallel Tempering (PT)}\label{alg:pt}

\begin{algorithmic}[1]
\REQUIRE{An objective function $h(x)$ to be minimized, or
equivalently the target distribution $\pi_{C} \ \alpha \
e^{-h(x)/C}$ for some positive constant $C$} \REQUIRE Swap
probability parameter $\alpha_0$
\REQUIRE{A proposal distribution $q(y|x)$}\\
\REQUIRE{Temperatures $T_1,T_2, \ldots, T_L$}, and number of
iterations $N$ \STATE Initialize arbitrary configurations
$x^{(1,1)},...,x^{(L,1)}($ \COMMENT{$x^{(k, i)}$: represents the
$i^{th}$ value of $x$ for temperature $T_k$;}) \STATE $x_{opt}
\leftarrow \argmin_{x=x^{(\cdot, 1)}}h(x)$
 \FOR{$i=1$ to $N$}
             \FOR{$k = 1$ to $L$}
                    \STATE $\widetilde{y} \leftarrow x^{(k, i-1)}$
                    \STATE Sample $y$ from the proposal distribution $q(y|\widetilde{y})$
                    \STATE $r \leftarrow
                    \min\{1,\frac{\pi_{T_k}(y)q(y|\widetilde{y})}{\pi_{T_k}(\widetilde{y})q(\widetilde{y}|y)}\}$
                    (Metropolis Hastings)
                    \STATE Draw $u \sim$ Uniform[0,1] and update
                      \IF{$u \leq r(\widetilde{y}, y)$}
                         \STATE $x^{(k, i)} \leftarrow y$
                       \ELSE \STATE $x^{(k, i)} \leftarrow \widetilde{y}$
                       \ENDIF
                   \IF{$h(x_{opt}) > h(x^{(k, i)})$}
                        \STATE $x_{opt} \leftarrow x^{(k, i)}$
                   \ENDIF
            \ENDFOR
            \STATE Draw $u \sim$ Uniform[0,1]
            \IF {$u\geq\alpha_0$}
                \STATE Draw $a$ Uniform $\{1,...,L-1\}$ and let
                $b=a+1$
                \STATE $r \leftarrow \min\{1,\frac{\pi_{T_a}(x^{(b,i)})\pi_{T_b}(x^{(a,i)})}{\pi_{T_a}(x^{(a,i)})\pi_{T_b}(x^{(b,i)})}\}$
                \STATE Draw $v \sim$ Uniform[0,1]
                \IF {$v\leq r$}
                    \STATE Swap $x^{(a,i)}$ and $x^{(b,i)}$
                \ENDIF
            \ENDIF
\ENDFOR
\end{algorithmic}
{\bf Return} $x_{opt}$
\end{algorithm}

If its swapping phase is excluded, PT is simply the combination of a fixed number of
Metropolis-Hastings procedures. The central point that makes PT powerful is its swapping step
where adjacent temperatures interchange their sampling regions. This
means that a good configuration can be allowed to use a cooler
temperature and exploit what it has found while a worse
configuration is given a higher temperature which results in more
exploration.

\section{The $\Phi$MDP Algorithm} \label{sec:agent}
We now describe how the generic $\Phi$MDP algorithm
works. The general algorithm is shown below (Algorithm
\ref{alg:phimdp}). It first takes a number of random actions ($5000$
in all our experiments). Then it defines the cost function
$Cost_{\alpha,\beta}$ based on this history. Stochastic search is then
used to find a map $\Phi$ with low cost. Based on the optimal $\Phi$ the history is transformed into a sequence of states, actions and rewards. We use optimistic frequency estimates from this history to estimate
probability parameters for state transitions and rewards. More precisely,
we use $\frac{R_{\max}+r_1+...+r_m}{m+1}$ instead of the average
$\frac{r_1+...+r_m}{m}$ to estimate expected reward, where $r_1,...,r_m$
are the rewards that have been observed for a certain state-action
pair, and $R_{\max}$ is the highest possible reward. The statistics
are used to estimate Q values using AVI. After this the agent starts
to interact with the environment again using $Q$-learning
initialized with the values that resulted from the performed AVI.
The switch from AVI to Q-Learning is rather obvious, as Q-Learning
only needs one cheap update per time step, while AVI requires updating the whole
environment model and running a number of value iterations. The first set of random actions might
not be sufficient to characterize what the best maps $\Phi$ look
like, so it might be beneficial to add the new history gathered by the Q-Learning interactions with the environment to the old history, and then
repeat the process but without the initial sampling.

\begin{algorithm}[!h]
\caption{Generic Stochastic $\Phi$MDP Agent (GS$\Phi$A)}
\label{alg:phimdp}
\begin{algorithmic}[1]
\REQUIRE $Environment$; $initialSampleNumber$, $agentLearningLoops$,
$stochasticIterations$ and  $additionalSampleNumber$ \STATE Generate
a history $h^{initial}$ of length $initialSampleNumber$ \STATE $h
\leftarrow h^{initial}$ \REPEAT
    \STATE Run the chosen stochastic search scheme for the history $h$ to find a $\hat{\Phi}$ with low cost
    \STATE Compute MDP statistics (optimistic frequency estimates $\hat{R}$ and $\hat{T}$) induced from $\hat{\Phi}$
    \STATE Apply AVI to find the optimal $Q^*$ values using the computed statistics
    $\hat{R}$ and $\hat{T}$.
    \STATE Interact with environment for $additionalSampleNumber$ iterations of Q-Learning using $Q^*$ as
    initial values; the obtained additional history is stored in $h^{additional}$
    \STATE $h \leftarrow [h, h^{additional}]$
    \STATE $agentLearningLoops \leftarrow agentLearningLoops - 1$
\UNTIL{$agentLearningLoops = 0$} \STATE Compute the optimal policy
$\pi^{optimal}$ from the optimal $\Phi$ and $Q$ values
\end{algorithmic}
{\bf Return} [$\Phi^{optimal}$, $\pi^{optimal}$]
\end{algorithm}

In the first four experiments in Section \ref{sec:exp}, PT is employed to search over the
$\Phi$ space of Markov AOCTs.



\subsection{Proposal Distribution for Stochastic Search over the Markov-AOCT Space}
The principal optional component of the above high-level algorithm,
GS$\Phi$A, is a stochastic search procedure of which some algorithms
have been presented in Section \ref{sec:stochasticsearch}. In these
algorithms, an essential technical detail is the proposal
distribution $q$. It is natural to generate the next tree (the next proposal or configuration) from the current tree by splitting or merging nodes. It is possible to express the exact form of our proposal distribution,
and based on this to explain how the next tree (next configuration)
is proposed from the current tree (current configuration). However,
the analytical form of the distribution is cumbersome to specify, so
for better exposition we opt to describe the exact
behavior of the tree proposal distribution instead.

 The stochastic search procedure starts with a Markov AOCT where all of the tree nodes are
 mergeable, and splittable. However, in the course of the search, a tree node might become
 unmergeable, but not the other way round; and a splittable node might turn to be unsplittable
 and vice versa. These specific transfering scenarios  are described as follows. A mergeable
 tree node of the current tree becomes unmergeable if the current tree is proposed from the
previous tree by splitting that node, and the cost of the current
tree is smaller than that of the previous tree. A splittable leaf
node of the current tree becomes unsplittable if the state
associated with that node is not present in the current history;
however, an unsplittable leaf node might revert to splittable when
the state associated with that node is present in the future updated
history. The constraint on merging is to keep good short-term memory
for predicting rewards, while the other on splitting is simply
following the Occam's razor principle.

\paradot{Merge and split permits}
Given some current tree at a particular point in time of the
stochastic search process, when considering the generation of the
next tree proposal, most of the tree nodes, though labeled
splittable and/or mergeable, might have no split, or
merge permit,  or neither. A node has split permit
if it is a leaf node with splittable label. When a leaf node has
been split, we simply add all possible children for this node, and
label the edges according to the definition of AOCTs. As
mentioned above, the newly added leaf nodes might be labeled
unmergeable if the cost of the new tree is smaller than that of the
old one; and these nodes might also be labeled unsplittable if the
states associated with the new leaf nodes are not present in the
current history. A node has merge permit if it is labeled
mergeable, and all of its children are leaf nodes. When a tree node
is merged, all the edges and nodes associated with its children are
removed.

\paradot{Markov-merge and Markov-split permits}
Since our search space is the class of Markov OACTs,
whenever a split or merge occurs, extra adjustments might be needed
to make the new tree Markov. After a split, there might be nodes
that make the tree violate the Markov assumption, and therefore, need to be
split. After we split all of those we have to check again to see if
any other nodes now need to be split. This goes on until we have a
Markov AOCT again. The same applies to merging.

\begin{figure}\centering
    \includegraphics[scale=0.7]{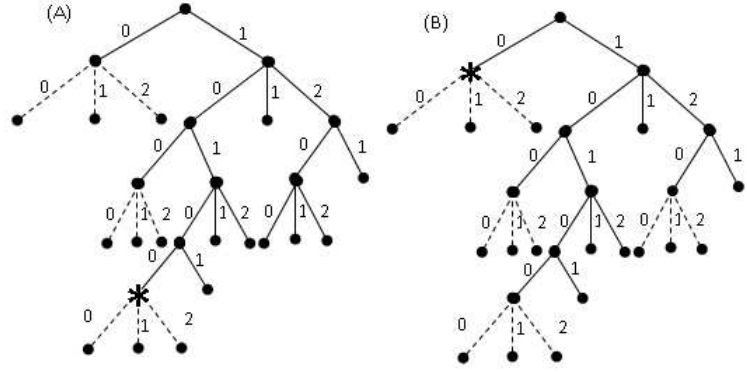}
    \caption{\textbf{AOCT proposals}}
    \label{fig:hybridcontexttreeproposals}
\end{figure}

When a node is Markov-split, it and all of the leaf nodes that need
to be split (including recursive splits) as a consequence in order
to make the tree Markov, are split. A tree node is said to have
Markov-split permit if it, and all the other nodes that would be split in a
Markov-split of the node, have split permits.
 This notion is best illustrated
with an example. First we define Markov and Non-Markov states of an
AOCT. A state of an AOCT is Markov if given any next
action-observation pair, the next state is determined; otherwise it
is labeled as non-Markov. Now in Figure
\ref{fig:hybridcontexttreeproposals}(A), suppose the current Markov AOCT is
the tree without dashed edges. Then after splitting the leaf node
marked by * (the node associated with state 00101), the state 001
becomes non-Markov so this associated node needs to be split.
However, after splitting this node (node associated with state 001),
state 0 becomes non-Markov, hence it needs splitting as well. In
short, to split the node marked by *, the two nodes associated with
states 001 and 0 have to be split as well so as to ensure the
resulting tree is Markov after splitting.  Similarly, a tree node
has Markov-merge permit if it, and all of the tree nodes that
minimally and recursively need to be merged after the original node
is merged in order to make the tree Markov, have merge permits. For
example, in Figure \ref{fig:hybridcontexttreeproposals}(B), suppose
the current tree is the tree including both solid and dashed edges,
then the node marked by * has Markov-merge permit, if it itself, and
the nodes associated with paths $001$, $021$ and $00101$ that need
to be merged, have merge permits. When a node with Markov-merge
permit is Markov-merged, it and its Markov-merge-associated nodes
are merged.

Our procedure to generate the next tree from the current tree (draw
sample from $q(y|\cdot)$) in the space of Markov AOCTs consists of the following main steps:

\begin{itemize}
    \item From the given tree, identify two sets: one is $N_S$ containing nodes with Markov-split permits,
    and the other $N_M$ containing nodes with Markov-merge permits.
    \item Suppose that either $N_S$ or $N_M$ is non-empty otherwise the algorithm (GS$\Phi$A) must stop;
    then if  either $N_S$ or $N_M$ is empty, select a node uniformly at random from the other set;
    otherwise select $N_S$ or $N_M$ randomly with probability $\frac{1}{2}$ each, and after that
    choose a tree node randomly from the selected set.
    \item Markov-split the node if it belongs to $N_S$, otherwise Markov-merge it
\end{itemize}
Once we have drawn the new tree $\tilde{y}$, the Metropolis Hastings
correction factor can be straightforwardly calculated via the formula
\begin{equation*}
\frac{q(y|\widetilde{y})}{q(\widetilde{y}|y)}=
\begin{cases}
    \frac{|\widetilde{N}_M|}{|N_S|}& \text{if}\ \widetilde{y}\ \text{is proposed from $y$ by Markov-splitting}\\
    \frac{|\widetilde{N}_S|}{|N_M|}& \text{if}\ \widetilde{y}\ \text{is proposed from $y$ by Markov-merging}
\end{cases}
\end{equation*}
here $\widetilde{N}_S$ and $\widetilde{N}_M$ are  respectively the set of nodes with Markov-split permits, and the set of nodes with Markov-merge permits of $\widetilde{y}$.

\paradot{Sharing}
If the stochastic search algorithm utilized is PT, we apply another trick to effectively accelerate the search process.
Whenever a node is labeled unmergeable, that is, by splitting this node the cost function decreases, or in other words a good additional relevant short-term memory for predicting rewards is found, the states associated
with the new nodes created by the splitting are replicated in the trees with the other
temperatures.

\section{Experiments}\label{sec:exp}

\subsection{Experimental Setup}

\begin{table}[!h]\centering
\begin{tabular}{c|c|c}
   \textbf{Parameter} & \textbf{Component} & \textbf{Value }\\
   \hline\hline
   $\alpha$& $Cost_{\alpha,\beta}$ & 0.1 \\
   $\beta$&  $Cost_{\alpha,\beta}$ & 0.1 \\
  $initialSampleNumber$ & GS$\Phi$A& 5000 \\
    $agentLearningLoops$ & GS$\Phi$A& 1 \\
   Iterations& PT & 100 \\
   $I$ & PT & 10 \\
   $T_i,\ i\leq I$& PT & $T_i = \beta \times i \times\log(n)$ \\
   $\alpha_0$& PT &  0.7\\
   $\gamma$ & AVI, Q-Learning & 0.999999\\
   $\eta$ & Q-Learning & 0.01\\
\end{tabular}
\label{tab:parametertable}
\vspace{3ex}
\caption{\textbf{Parameter setting for the GS$\Phi$A algorithm}}
\end{table}

Below in this section we present our empirical studies of the
$\Phi$MDP algorithm GS$\Phi$A described in Section \ref{sec:agent}. For all of our
experiments, stochastic search (PT) is applied in the $\Phi$ space of
Markov AOCTs.


For a variety of tested domains, our algorithm produces consistent
results using the same set of parameters. These parameters are shown in Table
\ref{tab:parametertable}, and are {\bf not} fine tuned.



The results of $\Phi$MDP and the three competitors in the four
above-listed environments are shown in Figures
\ref{fig:gridplot}, \ref{fig:tigerplot} \ref{fig:cheesemazeplot}, \ref{fig:Kuhnpokerplot}
and \ref{fig:relaymazeplot}. In each of the plots, various time
points are chosen to assess and compare the quality of the policies
learned by the four approaches. In order to evaluate how good a
learned policy is, at each point, the learning process of each agent, and the
exploration of the three competitors are temporarily
switched off. The selected statistic to compare the quality of
learning is the averaged reward over 5000 actions using the current
policy. For stability, the statistic is averaged over 10 runs.

As shown in more detail below, $\Phi$MDP is superior to U-tree and
active-LZ, and is comparable to MC-AIXI-CTW in short-term memory
domains. Overall conclusions are clear, and
we, therefore, omit error bars.



\subsection{Environments and results}
We describe each environment, the resulting performance, and the tree
that was found by $\Phi$MDP in the cheese maze domain.

\paradot{$4\times 4$  Grid}
\begin{figure}[!b]\centering
        \includegraphics[scale=0.5, angle=270]{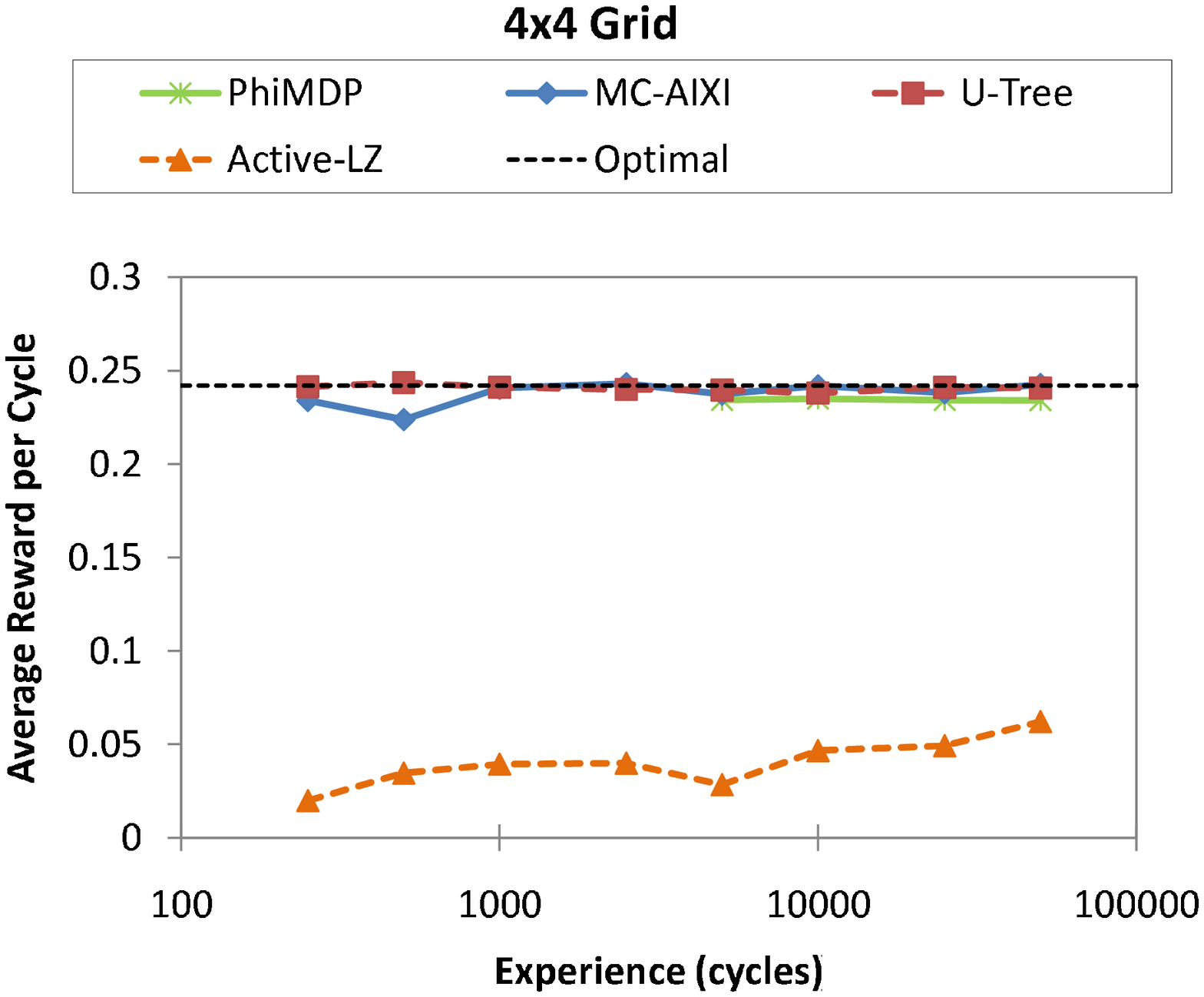}
         \caption{\textbf{$4\times4$ Grid}}
        \label{fig:gridplot}
\end{figure}
The domain is a 4$\times$4 grid world. At each time step, the agent can move one cell left,
right, up and down within the grid world. The observations are
uninformative. When the agent enters the bottom-right corner of the
grid; it gets a reward of 1, and is automatically and randomly sent
back to one of the remaining 15 cells. Entering any cell other than
the bottom-right one gives the agent a zero reward. To achieve the maximal total reward, the agent must be able to remember a series of smart actions without any clue about its relative position in the grid.

The context tree found contains 34 states. Some series of actions
that take the agent towards the bottom-right corner of the grid are
present in the context tree. As shown in the $4\times$4-grid plot in Figure \ref{fig:gridplot}, after 5000 experiences gathered from the random policy, $\Phi$MDP finds the optimal policy, and so does
MC-AIXI-CTW and U-Tree. Active-LZ, however, does not converge to an
optimal policy even after 50,000 learning cycles.

\paradot{Tiger}
The tiger domain is described as follows. There are two doors, left
and right; an amount of gold and a tiger are placed behind the two
doors in a random order. The person has three possible actions:
listen to predict the position of the tiger, open the right door,
and open the left door. If the person listens,  he has to pay some
money (reward of -1). The probability that the agent hears
correctly is 0.85. If the person opens either of the doors and sees
the gold, the obtained reward is 10; or otherwise he
faces the tiger, then the agent receives a reward of -100. After the
door is opened, the episode ends; and in the next episode the tiger
sits randomly again behind either the left or the right door.


\begin{figure}[!h]\centering
        \includegraphics[scale=0.5, angle=270]{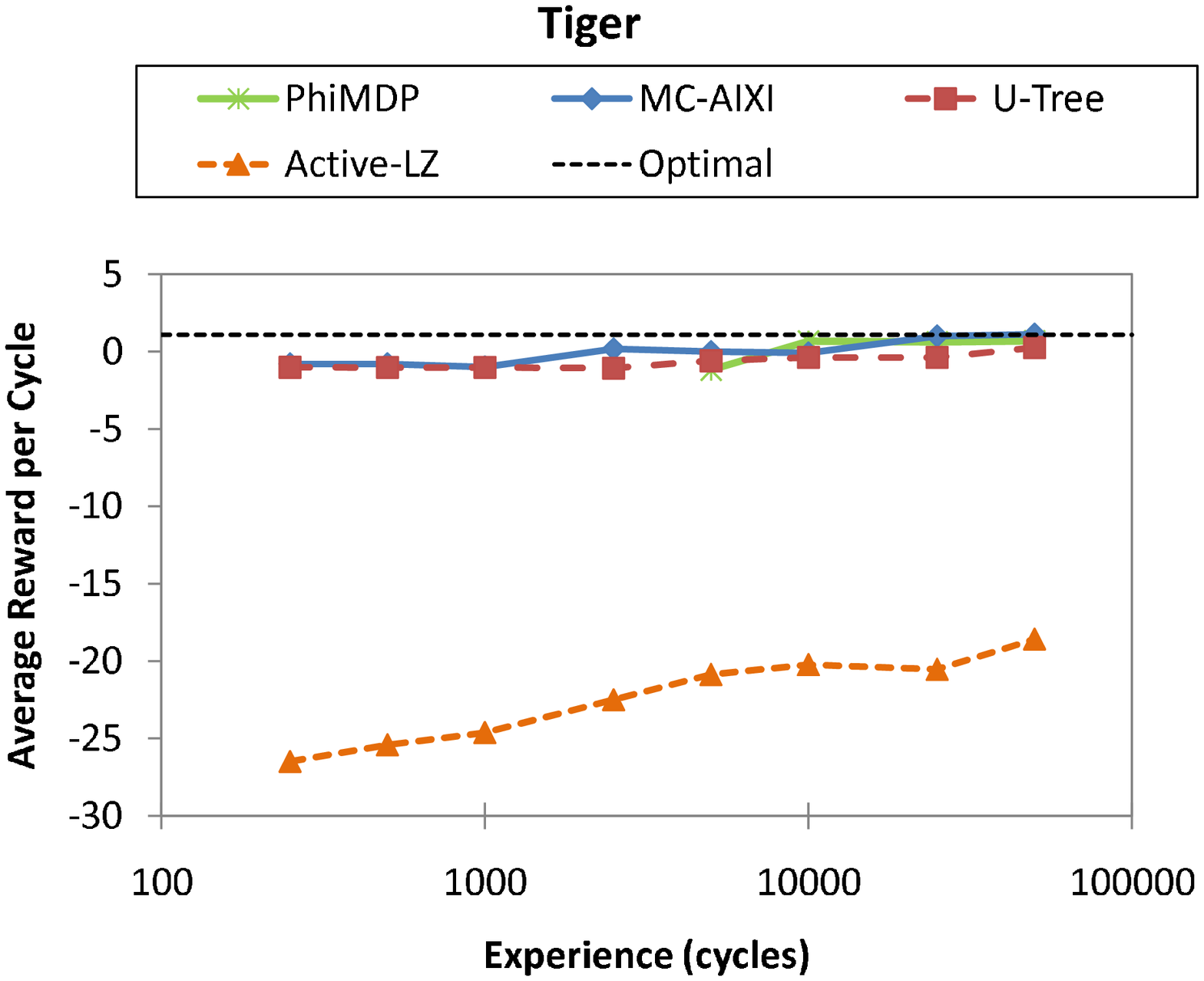}
    \caption{\textbf{Tiger}}
    \label{fig:tigerplot}
\end{figure}

Our parallel tempering procedure found a context tree consisting of
39 states including some important states where the history is such
that the agent has listened a few times before opening the door. It can be seen from the tiger plot in Figure \ref{fig:tigerplot} that the
optimal policy $\Phi$MDP found after 5,000 learning experiences does
yield positive reward on average, while from time point 10,000 on,
it achieves as high rewards as MC-AIXI-CTW. U-Tree appears to learn
more slowly but eventually manages to get positive averaged rewards
after 50,000 cycles like $\Phi$MDP and MC-AIXI-CTW. Active-LZ is
performing far worse. The optimal policy that $\Phi$MDP,
MC-AIXI-CTW, and U-Tree ultimately found is the following. First
listen two times, if the listening outcomes are consistent, open the
predicted door with gold behind; otherwise take one more listening
action, and based on the majority to open the appropriate door.

\paradot{Cheese Maze}
\begin{figure}[!h]\centering
  \includegraphics[scale=0.8]{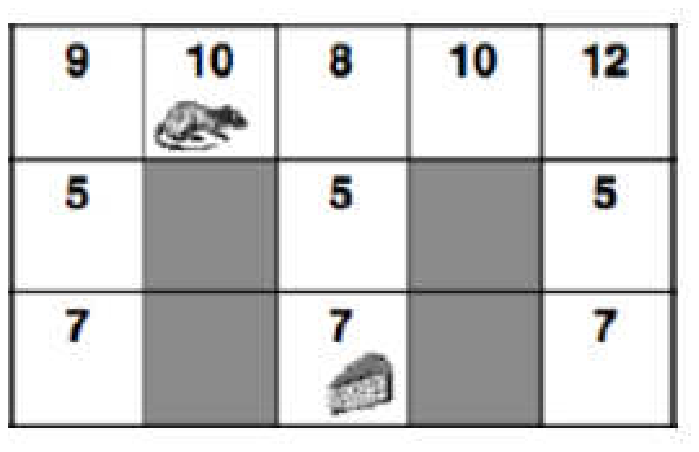} 
    \caption{\textbf{Cheese-maze domain}}
    \label{fig:cheesemazefigure}
\end{figure}
This domain, as shown in Figure \ref{fig:cheesemazefigure}, consists of a eleven-cell maze with a cheese in it.
The agent is a mouse that attempts to find the cheese. The agent's
starting position for each episode is at one of the eleven cells
uniformly random. The actions available to the agent are: move one
cell left (0), right (1), up (2) and down (3). However, it should be noticed that if the agent hits the wall,
its relative position in the maze remains unchanged. At each cell the agent can observe
which directions among left, right, up and down the cell is blocked
by a wall. If wall-blocking statuses of each cell are represented by 1 (blocked),
and 0 (free) respectively; then an observation is described by a
four-digit binary number where the digits from left to right are wall-blocking statuses
of up, left, down and right directions. For example, 0101 = 5, 0111
= 7, ... as described in Figure \ref{fig:cheesemazefigure}. The agent gets
a reward of -1 when moving into a free cell without a cheese;
hitting the wall gives it a penalty of -10; and a reward of 10 is
given to the agent when it finds the cheese. As can be seen, some observations themselves alone are insufficient for the mouse to locate itself unambiguously in the maze. Hence, the mouse must learn to resolve these ambiguities of observations in the maze to be able to find the optimal policy.

\begin{figure}[!h]\centering
    \includegraphics[scale=1.0]{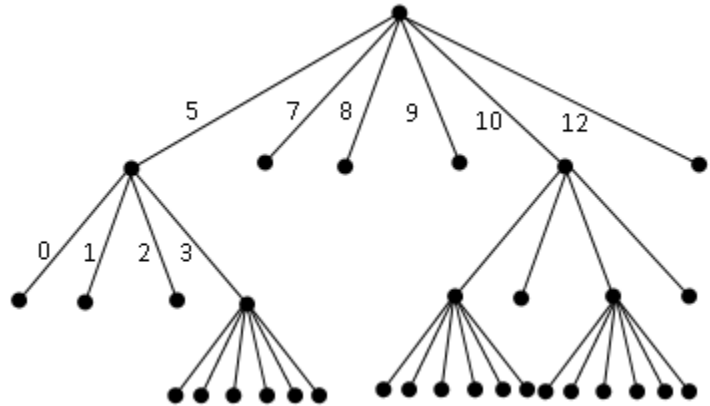}
    \caption{\textbf{Cheese-maze tree}}
    \label{fig:cheesemazetreee}
\end{figure}
\begin{figure}[!h]\centering
        \includegraphics[scale=0.5, angle=270]{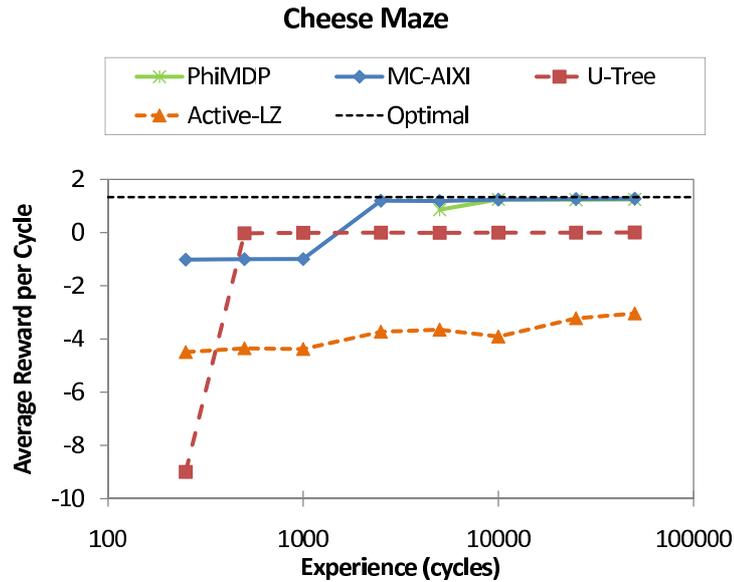}
            \caption{\textbf{Cheese maze}}
    \label{fig:cheesemazeplot}
\end{figure}

Our algorithm found a context tree consisting of 43 states that
contains the tree as shown in Figure \ref{fig:cheesemazetreee}. The
tree splits from the root into the $6$ possible observations. Then
observations $5$ and $10$ are split into the four possible actions;
and some of these actions, the ones that come from a different
location and not a wall collision, are split further into the $6$
``possible" observations before that. This resolves which $5$ or
which $10$ we are at. The states in this tree resolve the most
important ambiguities of the raw observations and an optimal policy
can be found. The domain contains an infinite amount of longer
dependencies among which our found states pick up a small subset.
The cheese-maze plot in Figure \ref{fig:cheesemazeplot} shows that after the initial 5000 experiences, $\Phi$MDP is marginally worse than MC-AIXI-CTW but is better than U-Tree and Active-LZ. From time
point 10,000, there is no difference between $\Phi$MDP and
MC-AIXI-CTW. U-Tree and Active-LZ remain inferior.


\paradot{Kuhn Poker}
\begin{figure}[!h]\centering
    \includegraphics[scale=0.5, angle=270]{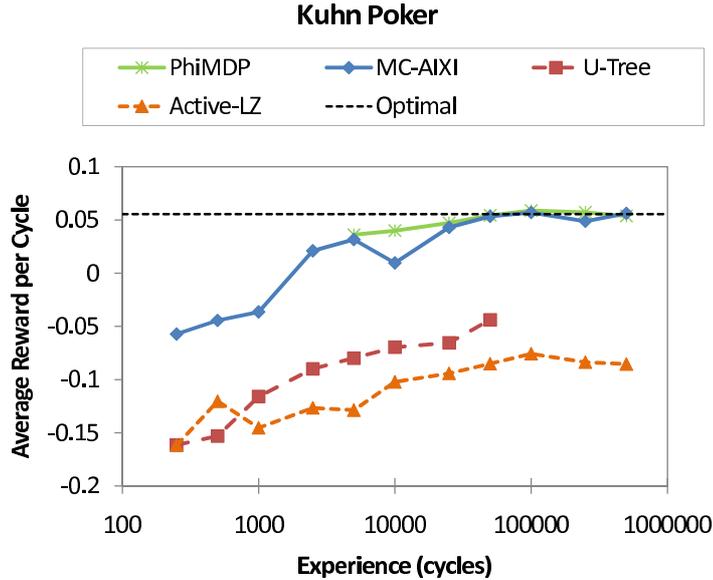}
    \caption{\textbf{Kuhn poker}}
    \label{fig:Kuhnpokerplot}
\end{figure}
In Kuhn poker \cite{HWK50}  a deck of only three cards (Jack, Queen
and King) is used.  The agent always plays second in any game
(episode). After putting a chip each into play, the players are
dealt a card each. Then the first player says bet or pass and the
second player chooses bet or pass. If player one says pass and
player two says bet then player one must choose again between bet
and pass. Whenever a player says bet they must put in another chip.
If one player bets and the other pass the better gets all the chips
in play. Otherwise the player with the highest card gets the chips.
Player one plays according to a fixed but stochastic Nash optimal
strategy \cite{HSHB05}. $\Phi$MDP finds $89$ states. It can  be observed from the Kunh-poker plot in Figure \ref{fig:Kuhnpokerplot} that $\Phi$MDP is
comparable to MC-AIXI-CTW and much better than U-Tree and Active-LZ, who loose money.
\section{Conclusions}\label{sec:concl}


Based on the Feature Reinforcement Learning framework \cite{MH09c}
we defined actual practical reinforcement learning agents that
perform very well empirically. We evaluated a reasonably simple
instantiation of our algorithm that first takes $5000$ random
actions followed by finding a map through a search procedure and
then it performs Q-learning on the MDP defined by the map's state
set.

We performed an evaluation on four test domains used to evaluate
MC-AIXI-CTW in \cite{VNHUS11}. Those domains are all suitably
attacked with context tree methods. We defined a $\Phi$MDP agent for
a class of maps based on context trees, and compared it to three other
context tree-based methods. Key to the success of our $\Phi$MDP
agent was the development of a suitable stochastic search method for
the class of Markov AOCTs. We combined parallel tempering with a
specialized proposal distribution that results in an effective
stochastic search procedure. The $\Phi$MDP agent outperforms both
the classical U-tree algorithm \cite{AKM96} and the recent Active-LZ
algorithm \cite{Far10}, and is competitive with the newest state of
the art method MC-AIXI-CTW \cite{VNHUS11}. The main reason that
$\Phi$MDP outperforms U-tree is that $\Phi$MDP uses a global
criterion (enabling the use of powerful global optimizers) whereas U-tree
uses a local split-merge criterion. $\Phi$MDP also performs
significantly better than Active-LZ. Active-LZ learns slowly as it
overestimates the environment model (assuming $n$-Markov or complete
context-tree environment models); and this leads to unreliable
value-function estimates.


Below are some detailed advantages of $\Phi$MDP over MC-AIXI-CTW:
\begin{itemize}
 \item $\Phi$MDP is more efficient than MC-AIXI-CTW in both computation and memory usage. $\Phi$MDP only needs
 an initial number of samples and then it finds the optimal map and uses AVI to
 find MDP parameters. After this it only needs a Q-learning update for each iteration.
 On the other hand, MC-AIXI-CTW requires model updating, planning and value-reverting at every
 single cycle which together are orders of magnitude more expensive than Q-learning. In the experiments
 $\Phi$MDP finished
 in minutes while MC-AIXI-CTW needed hours.
 Another disadvantage of MC-AIXI-CTW is that it is a memory-hungry algorithm. $\Phi$MDP learns
 the best tree representation using stochastic search, which expands a tree towards relevant histories.
 MC-AIXI-CTW learns the mixture of trees where the number of tree nodes grows
 (and thereby the memory usage) linearly with time.
 \item $\Phi$MDP learns a single state representation and can use many classical RL algorithms, e.g. Q-Learning, for MDP learning and planning.

\item Another key benefit  is that $\Phi$MDP represents a more discriminative approach than
MC-AIXI-CTW since it aims primarily for the ability to predict
future rewards and not to fully model the observation sequence. If
the observation sequence is very complex, this becomes essential.
\end{itemize}

On the other hand, to be fair it should be noted that compared to $\Phi$MDP, MC-AIXI-CTW is more principled.
The results presented in this paper are encouraging since they show that we can
achieve comparable results to the more sophisticated MC-AIXI-CTW
algorithm on problems where only short-term memory is needed. We plan to utilize
the aforementioned advantages of the $\Phi$MDP framework, like flexibility in environment modeling
and computational efficiency, to attack more complex and larger problems.

\section*{Acknowledgement}
This work was supported by ARC grant DP0988049 and by NICTA. We also thank  Joel Veness and Daniel Visentin for their assistance with the experimental comparison.

\bibliography{UIIA_References}

\bibliographystyle{splncs03}

\end{document}